\documentclass[aps,pre,preprint,superscriptaddress]{revtex4-1}

\usepackage{graphicx}
\usepackage{amssymb}
\usepackage{epsfig}
\usepackage{amsmath}
\usepackage{mathrsfs}
\usepackage{times}
\usepackage{bm}
\usepackage{color}
\usepackage{array}
\usepackage{makecell}
\begin{document}

\title{Transfer learning of chaotic systems}

\author{Yali Guo}
\affiliation{School of Physics and Information Technology, Shaanxi Normal University, Xi'an 710062, China}

\author{Han Zhang}
\affiliation{School of Physics and Information Technology, Shaanxi Normal University, Xi'an 710062, China}

\author{Liang Wang}
\affiliation{School of Physics and Information Technology, Shaanxi Normal University, Xi'an 710062, China}

\author{Huawei Fan}
\affiliation{School of Physics and Information Technology, Shaanxi Normal University, Xi'an 710062, China}

\author{Jinghua Xiao}
\affiliation{School of Science, Beijing University of Posts and Telecommunications, Beijing 100876, People’s Republic of China}

\author{Xingang Wang}
\email[Email address: ]{wangxg@snnu.edu.cn}
\affiliation{School of Physics and Information Technology, Shaanxi Normal University, Xi'an 710062, China}

\date{\today}

\begin{abstract} 
Can a neural network trained by the time series of system $\mathcal{A}$ be used to predict the evolution of system $\mathcal{B}$? This problem, knowing as transfer learning in a broad sense, is of great importance in machine learning and data mining, yet has not been addressed for chaotic systems. Here we investigate transfer learning of chaotic systems from the perspective of synchronization-based state inference, in which a reservoir computer trained by chaotic system $\mathcal{A}$ is used to infer the unmeasured variables of chaotic system $\mathcal{B}$, while $\mathcal{A}$ is different from $\mathcal{B}$ in either parameter or dynamics. It is found that if systems $\mathcal{A}$ and $\mathcal{B}$ are different in parameter, the reservoir computer can be well synchronized to system $\mathcal{B}$. However, if systems $\mathcal{A}$ and $\mathcal{B}$ are different in dynamics, the reservoir computer fails to synchronize with system $\mathcal{B}$ in general. Knowledge transfer along a chain of coupled reservoir computers is also studied, and it is found that, although the reservoir computers are trained by different systems, the unmeasured variables of the driving system can be successfully inferred by the remote reservoir computer. Finally, by an experiment of chaotic pendulum, we show that the knowledge learned from the modeling system can be used to predict the evolution of the experimental system.
\end{abstract}

\maketitle

{\bf A distinct difference between human brain and artificial neural network is that the former is able to apply the knowledge learned from one task to many other tasks, while the latter needs to lean from scratch for each task. This capability, knowing as transfer learning or knowledge transfer, is regarded as one of the key components for developing the next-generation artificial intelligence, and has spurred a new surge of wave in machine learning. By the technique of reservoir computer, we investigate in the present work the feasibility of transferring knowledge between two different dynamical systems. Our studies show that, given that the two systems are of the same type of dynamics, the reservoir trained by one system can be successfully adapted to predict the evolution of the other system, despite the clear difference between their motions. In particular, it is found that the reservoir computer trained by a periodic oscillator is able to infer the time evolution of a chaotic oscillator and even a spatiotemporal chaotic system. The feasibility of knowledge transfers is verified in a variety of modeling systems, as well as in an experimental system of chaotic pendulum. The findings shed lights on the working mechanism of reservoir computer, and facilitate the application of reservoir computer in practice.}   


\section{Introduction}

Model-free prediction of the evolution of chaotic systems by machine learning techniques has received many research interest in recent years~\cite{MTM:2002,ML:2009,Tanaka:2019,FocusIssue:Chaos2020}. Among the proposed techniques, reservoir computer (RC) is distinguished from others in stability, simplicity and efficiency, and has been employed as the main paradigm in literature for predicting the time evolution of complex dynamical systems~\cite{RC:Jaeger,Parlitz:2018,Ott:2018PRL,AW:2020}. From the perspective of dynamical systems, RC can be regarded as a complex network of coupled nonlinear elements which, driven by the input signals, generate the output data through a readout function. Except the parameters of the readout function, all other parameters and the dynamics of RC are fixed at the construction. In implementing RC, one first needs to set properly the readout parameters by a training process. After that, the system is closed by using the outputs as the driving signals, and then is evolved as an autonomous dynamical system, with the outputs being used for predictions. Although conceptually simple, RC has shown its power in a variety of applications, particularly in predicting and emulating chaotic systems~\cite{RC:Jaeger,Parlitz:2018,Ott:2018PRL,AW:2020}. 

Synchronization plays a crucial role in exploring and exploiting RC~\cite{LZX:2017,DIS:2018,Small:2019,Lmburn:2019,FHW:2020}. In predicting chaotic systems by the RC technique, a necessary condition is that the high-dimensional state of the reservoir network and the low-dimensional state of the modeling system should be related by a stable function, i.e., the two systems reach generalized synchronization~\cite{Rulkov:1995,Henry:1996,LZX:2018}. When generalized synchronization is established, not only the short-term behaviors of the chaotic systems can be forecasted, but also their long-term ergodic properties, such as Lyapunov exponents and attractors~\cite{Pathak:2017}, can be replicated. In exploiting RC for predicting the evolution of chaotic systems, a novel synchronization-based scheme has been proposed for inferring the unmeasured variables~\cite{LZX:2017}. In this scheme, the RC is firstly trained by the modeling system for an initial period and then driven by the latter through some measured variables. Interestingly, it is found that after a short transient period, the outputs of the RC become identical to that of the modeling system, i.e., the RC is completely synchronized with the modeling system, realizing thus the prediction of the unmeasured variables for an arbitrarily long horizon.    

In synchronizing RC with the modeling system, a general requirement in previous studies is that the training and driving signals should be generated by the same modeling system~\cite{LZX:2017,DIS:2018,Small:2019,Lmburn:2019,FHW:2020}. This requirement is natural from the point of view of synchronization~\cite{Book:PRK,SYNREV:Boccaletti}, as the purpose of the training process is to emulate the modeling system as close as possible, and identical systems are favorable for synchronization. In practice, however, this requirement might encounter some difficulties. For instance, in many realistic situations the training data are rare or, in some circumstances, expensive to obtain. In such a case, a great desire is that the RC trained by system $\mathcal{A}$, for which sufficient training data are available, can be used to predict the evolution of system $\mathcal{B}$, for which very few or even no training data are available. This question, knowing as transfer learning or knowledge transfer~\cite{SJP:2010,ZFZ:2015}, has emerged as a new learning framework in machine learning and attracted many attention in recent years. 

Employing the concept of transfer learning, in the present work we study the transfer of knowledge between different chaotic systems. The typical question we ask is the following: if a RC is trained by a short time series acquired from a periodic oscillator $\mathcal{A}$, can we use the trained RC to predict the evolution of the unmeasured variables of system $\mathcal{B}$ which is chaotic? Our main finding is that, given systems $\mathcal{A}$ and $\mathcal{B}$ have the same type of dynamics, the knowledge that RC learned from $\mathcal{A}$ can be transferred and be used to predict the evolution of $\mathcal{B}$; however, if the two systems are of different types of dynamics, transfer learning fails in general. We support our finding by demonstrating knowledge transfer in different modeling systems, as well as in an experimental system of chaotic pendulum. 

\section{Method}

We use the discrete-time echo state networks (ESN) to realize the RC~\cite{ML:2009}. Following Ref.~\cite{RC:Jaeger}, we construct the ESN by three modules: an input layer, a reservoir network, and an output layer. The input layer is described by the matrix $\mathbf{W}_{in}\in \mathbb{R}^{N\times (N_u+1)}$, with $N_u$ the dimension of the input vector $\mathbf{u}$. The elements of $\mathbf{W}_{in}$ is drawn randomly from the interval $\left[-\sigma,\sigma\right]$. The reservoir network consists of $N$ nonlinear nodes, which are connected in a random fashion with the probability $p$. The connections are characterized by the weighted matrix $\mathbf{A}\in \mathbb{R}^{N\times N}$, with $a_{ij}$ the strength of the connection pointing from node $j$ to node $i$. The value of $a_{ij}$ is drawn randomly within the range $\left[0,\eta\right]$, and is normalized by the largest eigenvalue of $\mathbf{A}$.  Once the ESN is constructed, the matrices $\mathbf{A}$ and $\mathbf{W}_{in}$ will be fixed. Driving by the input vector, the state of the reservoir network, $\mathbf{r}\in\mathbb{R}^N$, is updated according to the equation
\begin{equation}
\mathbf{r}(n+1)=(1-\alpha)\mathbf{r}(n)+\alpha\tanh\left\{\mathbf{Ar}(n)+\mathbf{W}_{in}\left[b_{in};\mathbf{u}(n)\right]\right\}.
\label{esn}
\end{equation}   
Here $\alpha\in(0,1]$ is the leaking rate and $b_{in}=1$ is the input bias. The initial conditions of the nodes in the reservoir are chosen randomly within the interval $(-1,1)$ and, before the training process, the ESN is evolved for a period $\tau$ in order of eliminating the transient states. 

The output layer is described by the function
\begin{equation}
\mathbf{v}(n)=\mathbf{W}_{out}\left[b_{out};\mathbf{u}(n-1);\mathbf{r}(n)\right],
\label{output}
\end{equation}
with $\mathbf{v}\in \mathbb{R}^{N_v}$ the output vector, $\mathbf{W}_{out}\in \mathbb{R}^{N_v\times (N+N_u+1)}$ the output matrix, and $b_{out}=1$ the output bias. For simplicity, we set $N_u=N_v$ in the present work. Unlike $\mathbf{A}$ and $\mathbf{W}_{in}$, the elements of $\mathbf{W}_{out}$ are not given at the construction, but to be ``learned" from the input signals through a training process. The purpose of the training is to find the proper elements for $\mathbf{W}_{out}$ such that the output vector $\mathbf{v}(n)$ is as close as possible to $\mathbf{u}(n)$ for $n=\tau+1,\ldots,\tau+T$. This can be done by minimizing a cost function with respect to $\mathbf{W}_{out}$~\cite{RC:Jaeger,Parlitz:2018,Ott:2018PRL,AW:2020}, which gives $\mathbf{W}_{out}=\mathbf{V}\mathbf{U}^T(\mathbf{U}\mathbf{U}^T+\lambda \mathbb{I})^{-1}$. Here, $\mathbf{U}\in \mathbb{R}^{(N+N_u+1)\times T}$ is the state matrix whose $k$th column is $\left[b_{out};\mathbf{u}(k);\mathbf{r}(k+1)\right]$, $\mathbf{V}\in \mathbb{R}^{N_v\times T}$ is a matrix whose $k$th column is $\mathbf{u}(k+1)$, $\mathbb{I}$ is the identity matrix, and $\lambda$ is the ridge regression parameter. After training, the matrix $\mathbf{W}_{out}$ is fixed in Eq. (\ref{output}), and the system is evolving autonomously by replacing $\mathbf{u}(n)$ with $\mathbf{v}(n)$ in Eq. (\ref{esn}).   

We adopt the classical Lorenz oscillator as the main model to demonstrate the RC-based chaos prediction and knowledge transfer. The dynamics of the Lorenz oscillator is described by equations $(dx/dt, dy/dt, dz/dt)^T=\left [a(y-x), \rho x-y-xz, cz+xy\right ]^T$. In our studies, we fix the parameters $a=10$ and $c=8/3$, and tune $\rho$ to change the system motion. The initial conditions of the oscillator are chosen randomly within the range $(-1,1)$, and the equations are solved numerically by the fourth-order Runger-Kutta algorithm, with the time step being $\delta t=0.02$. After a transient period of $t=1\times 10^3$, we start to record the system state, and $\tilde{T}=1\times 10^4$ time steps are recorded in total. The state variables are normalized to be within the range $\left[-1, 1\right]$. Among the processed data, the first $\tau=400$ points are used for eliminating the transient states of the reservoir, the next $T=2600$ points are used for training the output matrix, and the remaining points are used for testing purposes.  

\begin{figure}[tbp]
\center
\includegraphics[width=0.8\linewidth]{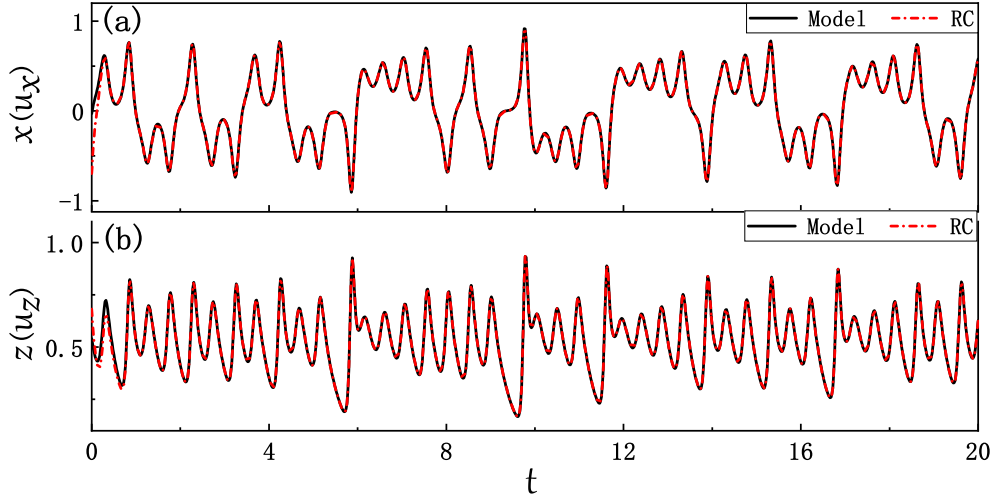}
\caption{Driven by the measurable variable $y$ of the chaotic Lorenz model, the RC outputs $u_x$ and $u_y$ are synchronized with the unmeasured variables $x$ and $y$ of the model, respectively. The training data and driving signals are generated by the same chaotic oscillator. $x$ and $z$ (black, solid lines): the unmeasured variables of the modeling system. $u_x$ and $u_y$ (red, dashed lines): the RC outputs.}
\label{fig1}
\end{figure}

Following Ref.~\cite{LZX:2017}, we adopt the drive-response scheme to realize the synchronization between the modeling system and RC. Specifically, in the predicting phase, we replace $\mathbf{u}(n)=(u_x, u_y, u_z)$ with $(u_x, y, u_z)$ in Eq. (\ref{esn}), and investigate the synchronization between the RC outputs $(u_x, u_z)$ and the unmeasured variables, $(x, z)$, of the modeling system. Previous studies have shown that if the training and test data are generated from the same system, $(u_x, u_z)$ and $(x, z)$ can be well synchronized, realizing the long-term prediction of the chaotic systems~\cite{LZX:2017,Small:2019,FHW:2020}. (Synchronization can be also achieved by replacing $u_x$ with $x$ or, with some skills, by replacing $u_z$ with $z$.) To demonstrate the synchronization between RC and modeling system, we set $\rho=60$ for the Lorenz model (with which the Lorenz oscillator is chaotic) and use the time series it generates to train and drive the RC. The parameters of the RC are $(N, p, \eta, \alpha, \sigma, \lambda) = (500, 0.25, 0.99, 0.95, 1.0, 1\times 10^{-10})$, which are obtained by a grid search in the hyperparameter space. Figure \ref{fig1} shows the time evolution of $u_x$ and $u_z$ generated by RC in the predicting phase. We see that after a short transient period (due to the random initial conditions of the trained RC), the motions of $u_x$ and $u_z$ are well synchronized with $x$ and $z$, respectively. The question we are interested and going to address in the following is: what happens if the training and driving signals are from different systems, saying, for example, training RC by a periodic Lorenz oscillator ($\rho=166$) while still driving it with the chaotic Lorenz oscillator ($\rho=60$)? 

\section{Results}

\begin{figure}[tbp]
\center
\includegraphics[width=0.95\linewidth]{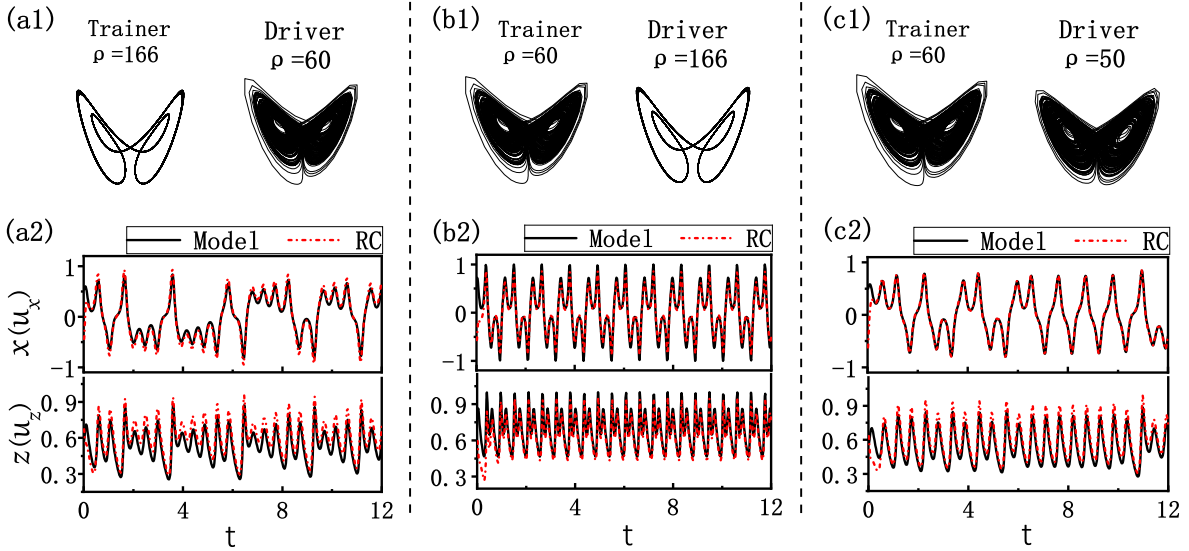}
\caption{Knowledge transfer between different Lorenz oscillators. (a) Training RC by period-$4$ oscillator ($\rho=166$) and driving the trained RC by chaotic oscillator ($\rho=60$). The parameters of RC are $(N, p, \eta, \alpha, \sigma, \lambda) = (500, 0.1, 0.99, 0.85, 1.0, 1\times 10^{-10})$. (b) Training RC by chaotic oscillator ($\rho=60$) and driving the trained RC by period-$4$ oscillator ($\rho=166$). (c) Training RC by chaotic oscillator ($\rho=60$) and driving the trained RC by another chaotic oscillator ($\rho=50$). Synchronization is observed between the driving system and the trained RC for the unmeasured variables in all three cases.}
\label{fig2}
\end{figure}

Comparing with chaotic signals, periodic signals are more convenient and efficient in training RC, especially in situations when only a small dataset is available, or when it is expensive to acquire new data. As such, in practical applications it is of great desire if the RC could be trained by periodic sequences while the trained RC can be used to predict chaotic sequences, namely transferring knowledge between different dynamical systems. Whereas in principle this is impossible if the trained RC is evolving autonomously in the predicting phase, but could be possible if in the predicting phase the RC is evolving non-autonomously, e.g., a coupling signal is received from the modeling system. To demonstrate, we set $\rho=166$ for the Lorenz oscillator [which generates period-$4$ motion, as depicted in Fig. \ref{fig2}(a1)], and use the signals it generated to train the RC. After the training, we reset the RC with random initial conditions and, as did in Fig. \ref{fig1}, drive it with the variable $y$ of the chaotic oscillator ($\rho=60$). The results are plotted in Fig. \ref{fig2}(a). We see that the outputs, $u_x$ and $u_z$, of the RC are well synchronized with the corresponding variables, $x$ and $z$, of the chaotic oscillator. We see that, indeed, the RC trained by the periodic Lorenz periodic is able to predict the evolution of chaotic Lorenz oscillator, i.e., the knowledge has been transferred.

Knowledge can also be transferred from chaotic to periodic systems, as well as between chaotic systems of the same type of dynamics but different parameters. The former is demonstrated in Fig. \ref{fig2}(b), in which the driving signal $y$ is generated by the period-$4$ Lorenz oscillator ($\rho=166$) and the RC is trained by the chaotic Lorenz oscillator ($\rho=60$). The latter is demonstrated in Fig. \ref{fig2}(c), in which the parameter of the driving system is chosen as $\rho=50$ (which generates chaotic motion), while the RC is trained by the chaotic oscillator with $\rho=60$. We see that in both cases the outputs of the RC are well synchronized with the corresponding variables of the driving system. 

\begin{figure}[tbp]
\center
\includegraphics[width=0.6\linewidth]{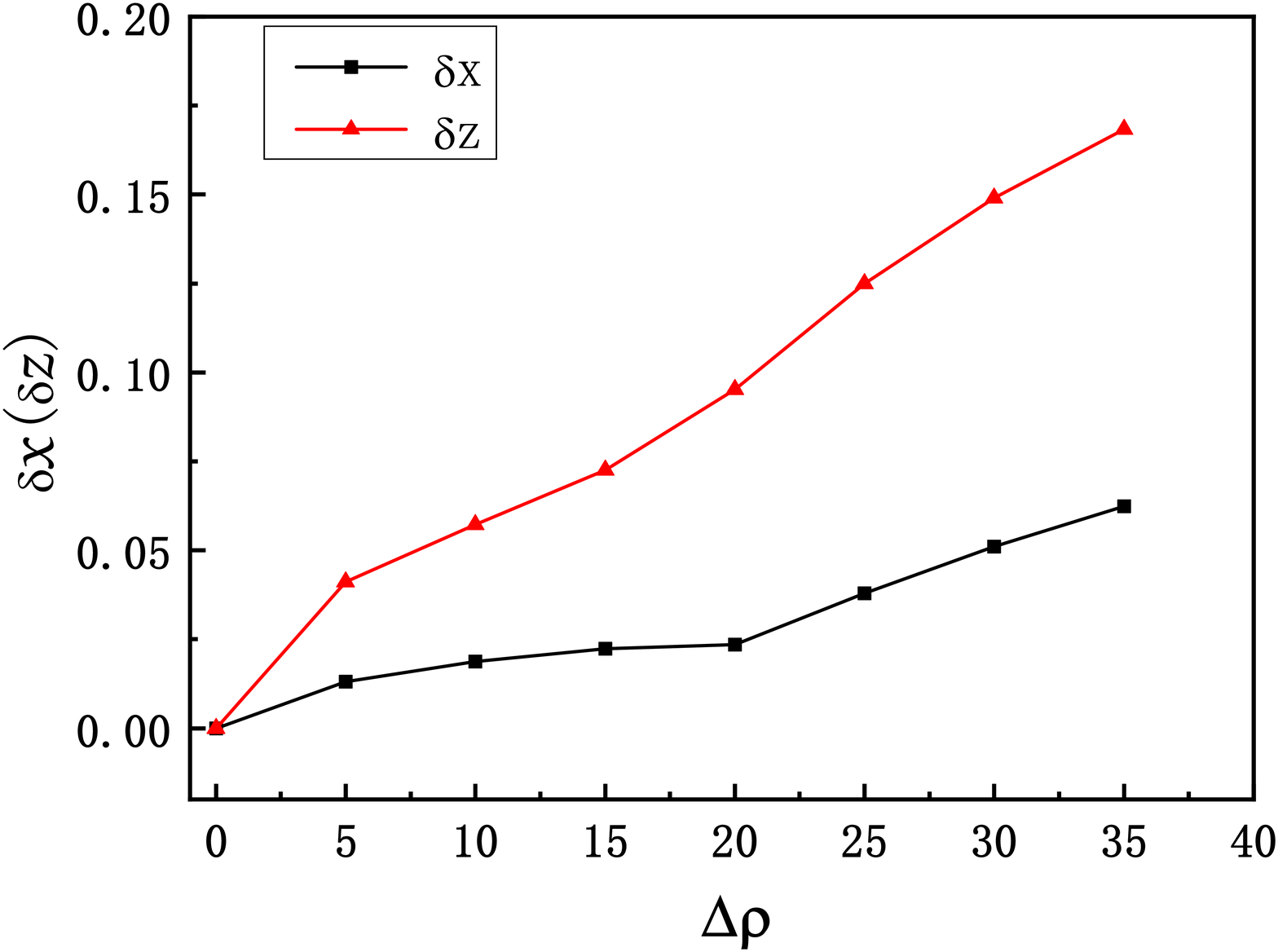}
\caption{Impact of parameter mismatch on knowledge transfer in Lorenz systems. The parameter of the training system is fixed as $\rho=60$ (chaotic motion), whereas the parameter of the driving system is $\rho=60-\Delta \rho$, with $\Delta \rho$ the parameter mismatch. Shown are the variations of the synchronization errors between the RC outputs and the driving system, $\delta x$ and $\delta z$, with respect to $\Delta \rho$.}
\label{fig3}
\end{figure}

Comparing with the results in Fig. \ref{fig1}, we see that in Fig. \ref{fig2} the predictions are slightly deteriorated. To be specific, in all three cases shown in Fig. \ref{fig2}, the output $u_z$ is synchronized in phase with the variable $z$, but with a noticeable difference in their amplitudes. This observation suggests that knowledge is not perfectly transferred between systems of different motions, making it interesting to study the impact of dynamics difference on the degree of knowledge transfer. To investigate, we fix the RC to be the one trained by the chaotic oscillator ($\rho=60$), and change the motion of the driving system by setting the parameter $\rho=60-\Delta \rho$, with $\Delta \rho$ the parameter mismatch quantifying the difference between the motions of the training and driving systems. The degree of knowledge transfer is measured by the time-averaged synchronization errors, $\delta x=\left<x-u_x\right>_T$ and $\delta z=\left<z-u_z\right>_T$,  with $\left< \ldots \right>_T$ the time-average function. The variations of $\delta x$ and $\delta z$ with respect to $\Delta \rho$ are plotted in Fig. \ref{fig3}. We see that as $\Delta \rho$ increases, both $\delta x$ and $\delta z$ are monotonically increased. These results are consistent with the findings in previous studies of oscillator synchronization, which show that synchronization is deteriorated in general as the parameter mismatch of the coupled systems increases~\cite{Book:PRK,SYNREV:Boccaletti}.  

\begin{figure}[tbp]
\center
\includegraphics[width=0.8\linewidth]{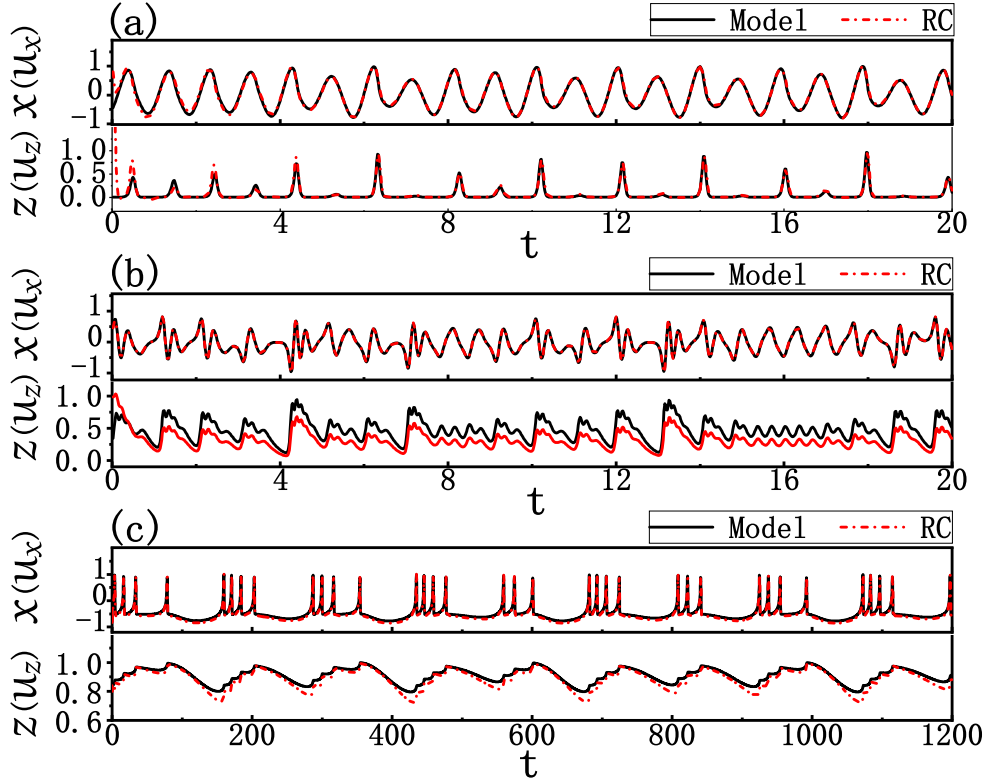}
\caption{Knowledge transfer in other types of chaotic systems. (a) R\"{o}ssler oscillator. Training system:  period-$4$ oscillator. Driving system: chaotic oscillator. The parameters of RC are $(N, p, \eta, \alpha, \sigma, \lambda) = (500, 0.27, 0.95, 0.95, 0.5, 1\times 10^{-10})$. (b) Chen oscillator. Training system:  period-$8$ oscillator. Driving system: chaotic oscillator. The parameters of RC are $(N, p, \eta, \alpha, \sigma, \lambda) = (500, 0.17, 0.8, 0.9, 0.65, 1\times 10^{-10})$. (c) Hindmarsh-Rose oscillator. Training system:  period-$3$ oscillator. Driving system: chaotic oscillator. The parameters of RC are $(N, p, \eta, \alpha, \sigma, \lambda) = (500, 0.36, 0.6, 0.8, 0.73, 8\times 10^{-6})$.}
\label{fig4}
\end{figure}

We proceed to check the generality of transfer learning in other types of chaotic systems, including R\"{o}ssler oscillators, Chen oscillators, and Hindmarsh-Rose (HR) oscillators. The dynamics of R\"{o}ssler oscillator is described by equations $(dx/dt, dy/dt, dz/dt)^T=\left [-y-z, x+ay, b+z(x-c)\right ]^T$. In our studies, we fix the parameters $a=b=0.2$, while adjusting the parameter $c$ to change the system dynamics. Specifically, the system presents the chaotic motion when $c=4.5$ and the period-$4$ motion when $c=4$. Training RC by the period-$4$ R\"{o}ssler oscillator, we drive the trained RC by variable $y$ of the chaotic R\"{o}ssler oscillator [replacing $u_y$ with $y$ in Eq. (1)]. The synchronization behavior between the RC outputs, $u_x$ and $u_z$, and the variables of the chaotic oscillator, $x$ and $y$, are plotted in Figure \ref{fig4}(a). We see that after a short transient period, $u_x$ is synchronized with $x$ and $u_z$ is synchronized with $z$. The dynamics of Chen oscillator is described by equations $(dx/dt, dy/dt, dz/dt)^T=\left [-a(y-x), (c-a)x+cy-xz, -bz+xy\right ]^T$. The system dynamics is chaotic for parameters $(a, b, c)=(35, 3, 28)$, and is periodic (period-$8$) for $(a, b, c)=(45, 3.18, 28)$. Still, we train RC by the period-$8$ Chen oscillator and drive the trained RC by the variable $y$ of the chaotic Chen oscillator. The results for Chen oscillator are presented in Fig. \ref{fig4}(b). We see that $u_x$ is well synchronized with $x$, while $u_z$ is slightly different from $z$ in amplitudes but is synchronized with $z$ in phase. The dynamics of HR oscillator is described by equations $(dx/dt, dy/dt, dz/dt)^T=\left [y-ax^3+bx^2-z+I, c-dx^2-y, r\left[s(x-x_0)-z\right] \right ]^T$. Fixing $(a, b, c, d, r, s, x_0)=(1, 3, 1, 5, 6\times 10^{-3}, 4, 1.56)$, the oscillator is chaotic for $I=2.8$ and is periodic (period-$3$) for $I=2.1$. Training RC by the period-$3$ HR oscillator and driving the trained RC by the chaotic HR oscillator, we plot in Fig. \ref{fig4}(c) the time evolutions of $u_{x}$, $x$, $u_{z}$, and $z$. We see that $u_x$ is synchronized with $x$ and $u_z$ is synchronized with $z$.  

\section{Generalization}

Can knowledge be transferred between systems with different types of dynamics? So far, our studies demonstrate only that knowledge can be transferred between systems with the same type of dynamics but different parameters, yet it remains not clear whether this can be done for systems with different types of dynamics. Whereas in principle it is impossible to transfer knowledge between oscillators of completely different dynamics, we find that under certain circumstances knowledge can still be transferred successfully. Specifically, we find that: (1) if the dynamical functions of the systems are correlated, e.g., by a linear transformation, knowledge can be transferred between the systems; (2) if the dynamics of the systems are completely different from each other, knowledge transfer fails in general, yet the driving system and the RC reach generalized synchronization.      

\begin{figure}[tbp]
\center
\includegraphics[width=0.8\linewidth]{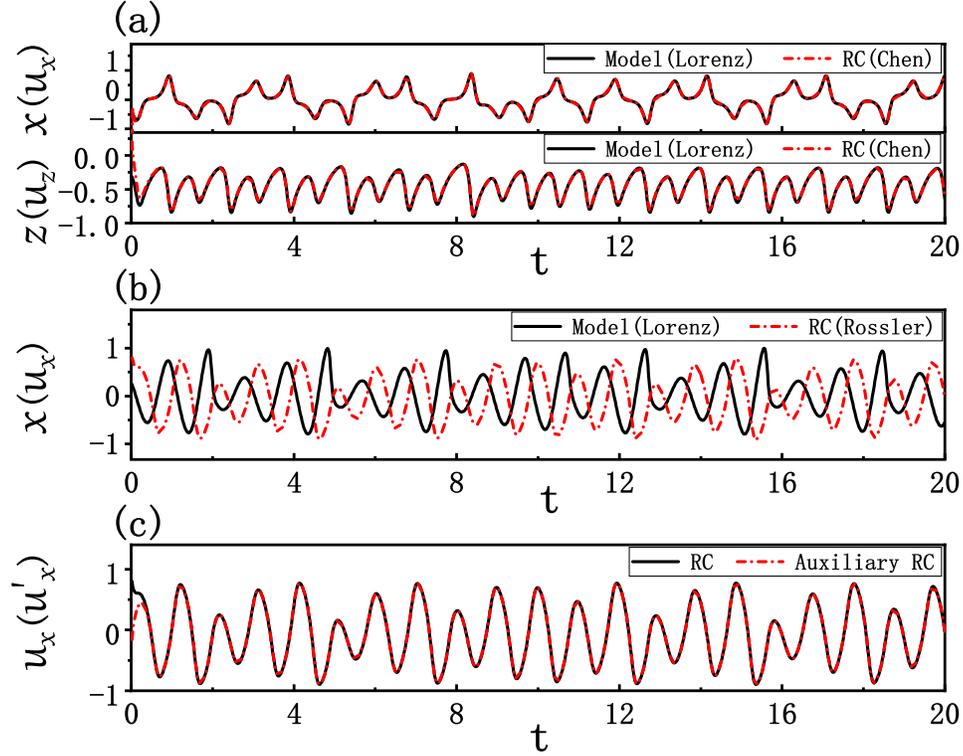}
\caption{Knowledge transfer between oscillators of different types of dynamics. (a) Knowledge transfer is realized between Chen and Lorenz oscillators. Training system: Chen oscillator of period-$8$ motion. Driving system: chaotic Lorenz oscillator. The parameters of the RC are $(N, p, \eta, \alpha, \sigma, \lambda) = (500, 0.25, 0.8, 0.5, 0.3, 1\times 10^{-9})$. (b) Knowledge transfer fails between R\"{o}ssler and Lorenz oscillators. Training system: chaotic R\"{o}ssler oscillator. Driving system: chaotic Lorenz oscillator. The parameters of the RC are $(N, p, \eta, \alpha, \sigma, \lambda) = (500, 0.2, 0.95, 0.9, 1.0, 1\times 10^{-10})$. (c) For RC trained by a chaotic R\"{o}ssler oscillator and driven by a chaotic Lorenz oscillator, complete synchronization is achieved between RC and its auxiliary, indicating that generalized synchronization is established between the RC and the driving system.}
\label{fig5}
\end{figure}

Knowledge can be transferred successfully between Lorenz and Chen oscillators. In Ref.~\cite{AA:2013}, it is argued that the Chen oscillator can be converted from the Lorenz oscillator by a homothetic transformation, i.e., the Chen oscillator can be regarded as a special case of the Lorenz oscillator. In Ref.~\cite{RB:20018}, it is argued that the two systems are distinct from each other in many features, but can be described by a unified system. Both studies suggest that the two types of oscillators, although of different attractors, share essentially the similar dynamics. Training RC by the period-$8$ Chen oscillator [as shown in Fig. \ref{fig4}(b)] and using the chaotic Lorenz oscillator [as shown in Fig. \ref{fig1}] as the driving system, we plot in Fig. \ref{fig5}(a) the time evolutions of the RC and the driving system. We see that, similar to the results shown in Fig. \ref{fig2}(a) (between two Lorenz oscillators) and Fig. \ref{fig4}(b) (between two Chen oscillators), the outputs $(u_x, u_z)$ of the RC are well synchronized with the corresponding variables $(x, z)$ of the driving system, justifying the feasibility of transferring knowledge between the two types of systems.

Knowledge transfer fails in general when the training and driving systems are of completely different dynamics. To show an example, we plot in Fig. \ref{fig5}(b) the synchronization behavior between the RC, which is trained by chaotic R\"{o}ssler oscillator, and the driving system represented by chaotic Lorenz oscillator. Still, the two systems are coupled by the variable $y$. Clearly, the RC output $u_x$ is desynchronized from the variable $x$, signifying the infeasibility of knowledge transfer between the two systems. The RC outputs and the chaotic Lorenz oscillator, however, are correlated in a generalized form, namely the generalized synchronization~\cite{Rulkov:1995}. This is shown in Fig. \ref{fig5}(c), in which the time evolutions of the RC output, $u_x$, and the output of the auxiliary RC, $u'_x$, are plotted. The auxiliary RC is identical to the RC trained by chaotic R\"{o}ssler oscillator and receives the same coupling signals, but is started from different initial conditions. Previous studies of oscillator synchronization show that~\cite{Henry:1996}, for generalized synchronization to be established between the driving and response systems, a necessary condition is that the response system loses its sensitivity to perturbations. For our case of knowledge transfer, this means that the outputs of the two RCs should be completely synchronized. This is indeed what we find in simulations. As shown in Fig. \ref{fig5}, the outputs of the RC and the auxiliary RC become identical after a short transient.    

\begin{figure}[tbp]
\center
\includegraphics[width=0.8\linewidth]{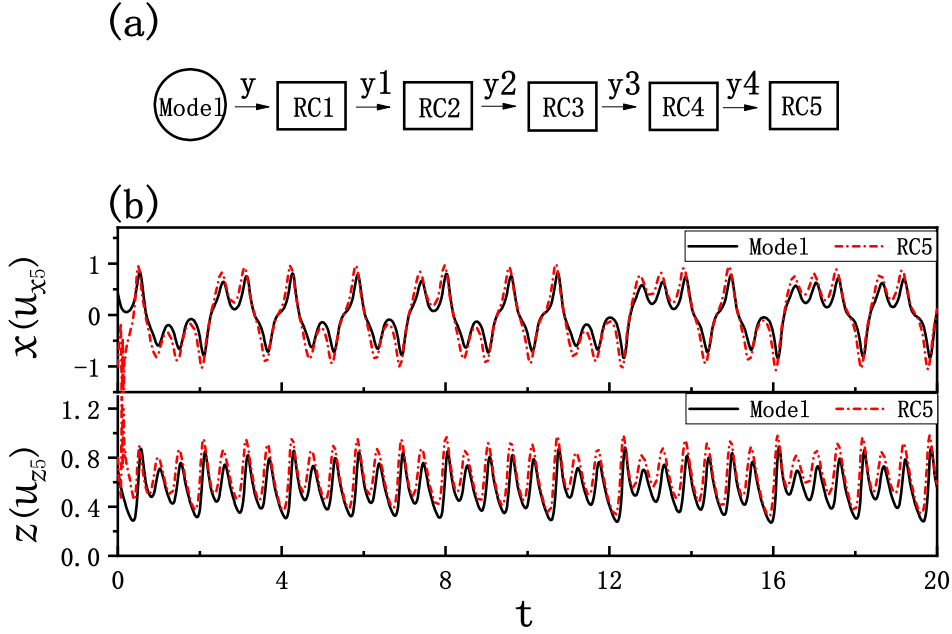}
\caption{Relaying knowledge along a chain of coupled RCs. (a) The coupling structure. Driving system (the left-most unit): chaotic Lorenz oscillator with $\rho=50$. RC1 is trained by chaotic Lorenz oscillator with $\rho=55$. The parameters of RC1 are $(N, p, \eta, \alpha, \sigma, \lambda) = (500, 0.35, 0.95, 0.8, 0.1, 1\times 10^{-10})$. RC2 is trained by period-$4$ Lorenz oscillator with $\rho=166$. RC3 is trained by chaotic Lorenz oscillator with $\rho=60$. RC4 is trained by chaotic Lorenz oscillator with $\rho=45$. Parameters of RC4 are $(N, p, \eta, \alpha, \sigma, \lambda) = (500, 0.25, 0.95, 0.9, 0.5, 1\times 10^{-10})$. RC5 is trained by period-$2$ Lorenz oscillator with $\rho=313$. Parameters of RC5 are $(N, p, \eta, \alpha, \sigma, \lambda) = (500, 0.15, 0.85, 0.99, 0.4, 1\times 10^{-11})$. (b) Synchronization behavior between the driving system and the outputs of RC5.}
\label{fig6}
\end{figure}

Can knowledge be transferred between RCs? In information engineering, a question of practical significance is whether the information acquired by the detecting system can be properly transferred to a remote system through a series of relay stations. If RCs are used as the relay stations, a natural question is whether information can be properly relayed among a series of RCs trained by different dynamical systems. To test, we design the hybrid system shown in Fig. \ref{fig6}(a), and check whether the unmeasured information of the driving oscillator (the left-most system) can be successfully inferred by the remote RC (the right-most system). It is worth noting that in the hybrid system, except the left-most system, all other systems are RCs and, importantly, the RCs are trained by different Lorenz oscillators (RC1 with $\rho=55$, RC2 with $\rho=166$, RC3 with $\rho=60$, RC4 with $\rho=45$, and RC5 with $\rho=313$). The time evolutions of the outputs, $u_{x5}$ and $u_{z5}$,  of the remote RC and the variables, $x$ and $z$, of the driving system are plotted in Fig. \ref{fig6}(b). We see that the remote RC is well synchronized with the driving system, indicating that the unmeasured variables as detected by the first RC are successfully relayed to the remote RC.       

How about high-dimensional spatiotemporal systems? Comparing with low-dimensional dynamical systems, the prediction of high-dimensional spatiotemporal systems using RC techniques is more challenging~\cite{ML:2009}. For instance, to emulate spatiotemporal dynamical systems by the RC technique, normally a huge reservoir should be constructed and a large amount of training data are acquired. Recent studies show that for some spatiotemporal systems, the difficulties could be solved by the scheme of parallel RCs~\cite{Parlitz:2018,Ott:2018PRL}. In the simplest version of this scheme, first a small-size RC is trained by the local dynamics of the system, then an ensemble of RCs (which has the same parameters as the trained RC) are coupled on a lattice to emulate the large-size system. It is shown that the parallel scheme, while reducing significantly the computational cost, yields excellent predictions for the evolution of typical spatiotemporal systems. Employing the parallel scheme, we next investigate knowledge transfer between two spatiotemporal chaotic systems.   

\begin{figure}[tbp]
\center
\includegraphics[width=0.8\linewidth]{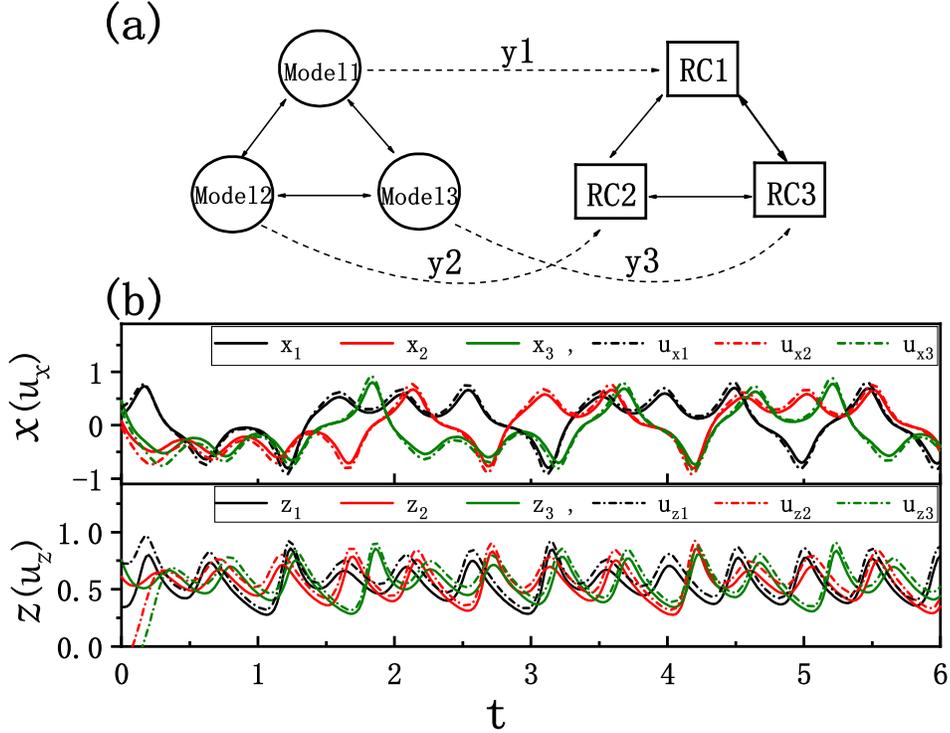}
\caption{Knowledge transfer between two spatiotemporal systems. (a) The modeling system (left) and the large-size RC (right). The driving system consists of three desynchronized chaotic oscillators. The large-size RC is constituted by three identical RCs, which are trained by the time series of synchronized periodic oscillators. The RCs are initialized with different initial conditions. (b) The synchronization behavior of the modeling system and the large-size RC. $x_{1,2,3}$ and $z_{1,2,3}$ are the variables of the modeling system; $u_{x1,x2,x3}$ and $u_{z1,z2,z3}$ are the outputs of the RCs.}
\label{fig7}
\end{figure}

Our model of spatiotemporal system consists of three globally coupled identical oscillators, as schematically plotted in the left side of Fig. \ref{fig7}(a). The dynamics of the $i$th oscillator is governed by equations $\dot{\mathbf{x}_i}=\mathbf{F}(\mathbf{x}_i)+\varepsilon\sum_j\left[\mathbf{H}(\mathbf{x}_j)-\mathbf{H}(\mathbf{x}_i)\right]$, with $\dot{\mathbf{x}}=\mathbf{F}(\mathbf{x})$ representing the dynamics of isolated oscillator, $\varepsilon$ denoting the uniform coupling strength, and $\mathbf{H}(\mathbf{x})$ is the coupling function. For illustration, we adopt the Lorenz oscillator as the local dynamics, and utilize the diagonal coupling function $\mathbf{H}(\mathbf{x})=[x,y,z]^T$. To generate the training signals, we set $\rho=166$ for the oscillators and choose the coupling strength $\varepsilon=1\times 10^{-3}$, by which the oscillators present the synchronized periodic motions. The parameters of the RC are $(N, p, \eta, \alpha, \sigma, \lambda) = (500, 0.35, 0.99, 0.95, 1.0, 1\times 10^{-10})$, and the RC is trained by the time series of the $1$st oscillator, $(x_1,y_1,z_1)$. We make two copies of the trained RC, and connect the three RCs to form a large-size RC, with the coupling structure and function identical to that of the modeling system [see Fig. \ref{fig7}(a)]. The question we are interested is whether the large-size RC, which is trained by the time series of synchronized periodic oscillators, is able to infer the time evolution of spatiotemporal chaos of desynchronized chaotic oscillators. To investigate, we construct the driving system by replacing the periodic oscillators with chaotic oscillators ($\rho=60$) in the model, and set the coupling strength as $\varepsilon=2\times 10^{-2}$. Analysis based on the method of master stability function shows that the chaotic oscillators are synchronized when $\varepsilon>\varepsilon_c\approx 0.47$~\cite{GS:Pecora,GS:Huang}. So, the chaotic oscillators are desynchronized from each other, and the driving system is spatiotemporally chaotic. Setting the coupling strength of the RCs as $\varepsilon=2\times 10^{-2}$, we drive the large-size RC by replacing $u_{yi}$ with $y_i$ ($i=1,2,3$) for each RC [see Fig. \ref{fig7}(a)], and investigate the synchronization behavior between the large-size RC and the modeling system. The results are presented in Fig. \ref{fig7}(b). We see that the unmeasured variables of the spatiotemporally chaotic system, $(x_i,z_i)$, are well synchronized with the outputs of the large-size RC, $(u_{xi},u_{zi})$, with $i=1,2,3$ the oscillator (RC) index. 

\section{Experimental Verification}

\begin{figure}[tbp]
\center
\includegraphics[width=0.85\linewidth]{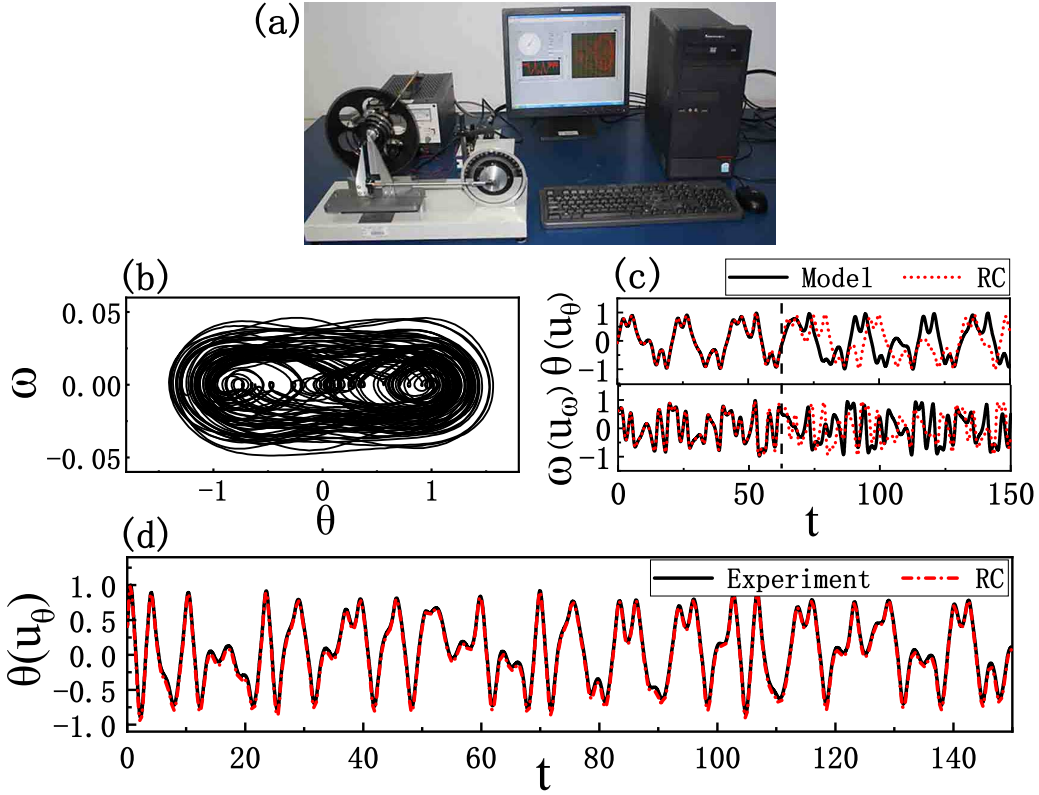}
\caption{Transferring knowledge between modeling and experimental systems. (a) The experimental setup. (b) The chaotic motion observed in experiment. (c) The performance of RC in predicting the time evolution of the modeling system. (d) The performance of RC in predicting the time evolution of the unmeasured variable $\theta(t)$ of the experimental pendulum. Black curve: the experimental results. Red curve: the results predicted by model-trained RC. }
\label{fig8}
\end{figure}

We finally check the feasibility of transferring knowledge between modeling and experimental systems. The system we adopt in experiment is the modified Pohl's torsion pendulum developed in Ref.~\cite{XJH:2014}. The experimental setup is shown in Fig. \ref{fig8}(a), in which a fine iron bar with a matching nut is added to the conventional apparatus Pohl’s torsion pendulum, and the pendulum is driven by a periodic force provided by the motor. Both the amplitude and frequency of the driving force are controlled by the computer. By varying the frequency of the driving force $\omega_d$, the system can be conveniently changed between the periodic and chaotic motions, with the trajectories displayed timely on the computer screen. The instant state of the pendulum is characterized by the swing angle $\theta(t)$, which is acquired by an angle sensor attached to the pivot. The sampling frequency is $f_c=50$Hz. The instant swing frequency is defined as $\omega(t)=[\theta(t+\delta t)-\theta(t)]/\delta t$, with $\delta t=1/f_c$. The data are processed by a low-pass filter, so as to smooth the trajectory. An example of chaotic motion is shown in Fig. \ref{fig8}(b), which is generated by the driving frequency $\omega_d=1.0$.  

It is straightforward to write the following equations for the motion of the pendulum, 
\begin{eqnarray}
\left\{
\begin{array}{lll}
&&\dot{\theta}=\omega,\\
&&\dot{\omega}=[-\gamma\omega-k\theta+mgr\sin(\theta)+M\cos(\omega_d t)]/J,
\end{array}
\right.
\label{expmodel}
\end{eqnarray}
with $J$ the moment inertia of the rotating wheel, $\gamma$ the friction coefficient, $k$ the stiffness coefficient of the spring, $m$ the mass of the iron bar and the nut, $g=9.8$ the gravity acceleration, $r$ the distance from the barycenter of the iron bar to wheel pivot, and $M$ the amplitude of the driving torque. If the above parameters can be precisely measured in experiment, we should be able to reproduce the chaotic attractor shown in Fig. \ref{fig8}(b) by solving Eq. (\ref{expmodel}) numerically. Yet, except $\omega_d$ and $M$ which are controlled by the computer, all other parameters are not known in our experiment. The question we are interested here is: if none of the experimental parameters (including $\omega_d$ and $M$) is known and, in the course of time evolution, only the information of the instant swing frequency, $\omega(t)$, is available, can we infer from $\omega(t)$ the time evolution of the instant swing angle, $\theta(t)$? 

While the above question can be addressed by many approaches, we tackle it by the approach of RC from the point of view of transfer learning. To implement, first we choose a set of parameters in the model capable of generating the typical chaotic motion. Then, we construct the RC and train it by the time series, $\theta(t)$ and $\omega(t)$, acquired from the chaotic model. The driving frequency $\omega_d$ is treated as an additional, constant input of the RC. The outputs of the RC are denoted as $u_{\theta}(t)$ and $u_{\omega}(t)$, which correspond to $\theta(t)$ and $\omega(t)$ in the modeling system, respectively. Finally, in the predicting phase, the RC is coupled to the experimental system by replacing $u_{\omega}(t)$ with $\omega(t)$, and the output $u_{\theta}(t)$ is used to predict the instant pendulum state $\theta(t)$. Setting $(J, \gamma, k, m, r, M, \omega_d)=(1.0,0.3,5,3,0.2,0.4,1.0)$ in Eq. (\ref{expmodel}), the modeling system presents the chaotic motion, with the largest Lyapunov exponent being $\Lambda\approx 0.24$. In simulations, the time step is chosen as $\delta t=0.05$, and, after a transient period, we collect $T=8\times 10^4$ states for the training data and an additional series of $T=1\times 10^3$ states as the test data. The parameters of RC are $(N, p, \eta, \alpha, \sigma, \lambda) = (500, 0.3, 0.8, 0.5, 1.0, 2\times 10^{-8})$. As depicted in Fig. \ref{fig8}(c), the trained RC is able to predict the time evolution of the modeling system for a period about $14$ Lyapunov times ($T\approx 60$). Replacing $u_w(t)$ with $w(t)$ (which is acquired from the experiment), we evolve the trained RC and plot in Fig. \ref{fig8}(d) the time evolution of the RC output $u_{\theta}(t)$. We see that the output $u_{\theta}(t)$ is identical to the unmeasured state $\theta(t)$ of the pendulum in the course of time evolution, signifying that the knowledge learned from the modeling system has been successfully transferred to the experimental system. 

\section{Discussions and conclusion}

We would like to note that the mission of transfer learning is using the knowledge learned from a particular task to other tasks that are related but different from it. More specifically, given a source domain $\mathcal{D}_S=\{\mathcal{X}_S,\mathit{P}_S(\mathcal{X})\}$ and learning task $\mathcal{T}_S$, and a target domain $\mathcal{D}_T=\{\mathcal{X}_T,\mathit{P}_T(\mathcal{X})\}$ and learning task $\mathcal{T}_T$, transfer learning aims to improving the learning of the target predictive function $\mathit{f}_T(\cdot)$ in $\mathcal{D}_T$ using the knowledge learned in $\mathcal{D}_S$ and $\mathcal{T}_S$, where $\mathcal{D}_S\neq \mathcal{D}_T$, or $\mathcal{T}_S\neq \mathcal{T}_T$~\cite{SJP:2010}. In general, the domain $\mathcal{D}$ consists of two components: a feature space $\mathcal{X}$ and a marginal probability distribution $\mathit{P}(\mathcal{X})$. For the question of knowledge transfer between different chaotic systems, the task in the training phase is to emulate the system dynamics, while in the predicting phase the task is to infer the unmeasured variables from the measured ones. The two tasks are correlated (in the sense that both require an accurate replication of the system dynamics) but different from each other (in the sense that in the predicting phase the RC is evolving autonomously in the former while is coupled to the driving system in the latter). Our studies show that under certain circumstances, the knowledge learned from the source domain, e.g., the time series of a periodic oscillator, can be used to infer the properties of the target domain, e.g., the time series of a chaotic oscillator. The different domains makes transfer learning distinguished from other tasks of RC such as chaos prediction and variable inference~\cite{RC:Jaeger,Parlitz:2018,Ott:2018PRL,AW:2020,LZX:2017,DIS:2018,Small:2019,Lmburn:2019,FHW:2020}, as in the latter the source and target domains are identical. 

The finding that knowledge can be transferred between systems of different dynamics sheds lights on the working mechanism of RC. Whereas general principles and criteria have been proposed for designing and training RC in literature~\cite{LZX:2018,LYC:2019,TLC:2019,TL:2019,AG:2019,AH:2019}, it remains not clear how RC functions, making RC essentially a ``black box" in applications. In particular, whereas RC is able to predict the time evolution of chaotic system with the unprecedented performance, it remains not clear what RC really learns from the training data -- the dynamical functions or the ergodic properties. If the knowledge is associated with the ergodic properties of the attractor, namely the climate~\cite{Pathak:2017}, then both the dynamical functions and the bifurcation parameters are critical for the learning. Otherwise, if the knowledge is associated with the dynamical functions, then the bifurcation parameters can be treated as an independent input through which the climate can be tuned flexibly. Our studies seem to suggest that, so far as the inference of unmeasured variables in chaotic system is concerned, the knowledge of the dynamical functions is learned separately from that of the bifurcation parameters. This finding is consistent with the results in recent studies of RC-based chaos predictions, where it is shown that the bifurcation diagram of chaotic systems can be obtained by treating the bifurcation parameters as an individual input~\cite{LYC:preprint}. 

To summarize, inspired by the recent studies of transfer learning in machine learning, we have exploited RC for transferring knowledge between different chaotic systems. It is found that for systems of the same type of dynamics, knowledge learned from one system can be used to infer the unmeasured variables of the other system, despite the difference of their parameters. However, when the systems are of different types of dynamics, knowledge transfer fails in general. Knowledge transfer in spatiotemporal systems of coupled oscillators and in a chain of coupled RCs has been also investigated, with the results supporting the findings obtained in low-dimensional systems. Finally, an experimental study has been conducted to verify the numerical findings, which show that knowledge learned from the modeling system can be successfully transferred to predicting the evolution of the unmeasured variable in experiment. The findings shed lights on the working mechanism of RC, and provide new perspectives for the applications of RC.\\

\noindent 
{\bf ACKNOWLEDGEMENTS}

This work was supported by the National Natural Science Foundation of China (NSFC) under the Grant No.~11875182.\\

\noindent 
{\bf DATA AVAILABILITY}

The data that support the findings of this study are available from the corresponding author
upon reasonable request.

\end{document}